\documentclass[sigconf]{acmart}
 
\usepackage{amsmath, amsfonts, amssymb, amsthm, xspace, color}
\usepackage{booktabs} 

\usepackage{subfigure, multirow, tabularx} 
\usepackage{booktabs} 
\usepackage{enumitem}
\usepackage{flushend}
\usepackage[T1]{fontenc}
\usepackage{epstopdf}
\usepackage{balance}
\usepackage{graphicx}
\usepackage[noend,ruled,linesnumbered]{algorithm2e} 
\usepackage{microtype}
\usepackage{url}
\usepackage{wrapfig,lipsum}
\usepackage{makecell}
\usepackage{wrapfig}
\usepackage{setspace}
\newcommand{\hide}[1]{} 

\newcommand{\nop}[1]{}

\newtheorem{thm:def}{Definition}
\newtheorem{thm:eg}{Example}
\newtheorem{thm:lem}{Lemma}
\newtheorem{thm:obs}{Observation}
\newtheorem{thm:req}{Requirement}
\newtheorem{thm:prop}{Proposition}
\newtheorem{thm:principle}{Principle}
\newtheorem{thm:thm}{Theorem}
\newtheorem{thm:corollary}{Corollary}

\def \T {\mathcal{T}}

\newcommand{\OurGCN}{\mbox{\sf BiTe-GCN}\xspace}
\newcommand{\OurGCNBf}{\mbox{\sf \textbf{BiTe-GCN}}\xspace}
\definecolor{midnightgreen}{rgb}{0.0, 0.29, 0.33}
\definecolor{orange}{RGB}{255,127,0}

\copyrightyear{2021}
\acmYear{2021}
\setcopyright{acmcopyright}\acmConference[WSDM '21]{Proceedings of the Fourteenth ACM International Conference on Web Search and Data Mining}{March 8--12, 2021}{Virtual Event, Israel}
\acmBooktitle{Proceedings of the Fourteenth ACM International Conference on Web Search and Data Mining (WSDM '21), March 8--12, 2021, Virtual Event, Israel}
\acmPrice{15.00}
\acmDOI{10.1145/3437963.3441774}
\acmISBN{978-1-4503-8297-7/21/03}

\begin{document}
\title{\OurGCNBf: A New GCN Architecture via Bidirectional Convolution of Topology and Features on Text-Rich Networks} 
\author{Di Jin$^{*1}$, Xiangchen Song$^{*2}$, Zhizhi Yu$^{1}$, Ziyang Liu$^{3}$, Heling Zhang$^{2}$, Zhaomeng Cheng$^{3}$, and Jiawei Han$^{2}$}
\affiliation{
\institution{$^1$College of Intelligence and Computing, Tianjin University, Tianjin, China} 
\institution{$^2$Department of Computer Science, University of Illinois at Urbana-Champaign, Champaign, IL, USA}
\institution{$^3$JD.com}
\institution{$^{1}$\{jindi, yuzhizhi\}@tju.edu.cn, $^2$\{xs22, hzhng120, hanj\}@illinois.edu, $^3$\{liuziyang7, chengzhaomeng\}@jd.com}
\country{}
}
\renewcommand{\shortauthors}{Jin and Song, et al.}
\begin{abstract}
Graph convolutional networks (GCNs), aiming to obtain node embeddings by integrating high-order neighborhood information through stacked graph convolution layers, have demonstrated great power in many network analysis tasks such as node classification and link prediction. However, a fundamental weakness of GCNs, that is, topological limitations, including over-smoothing and local homophily of topology, limits their ability to represent networks. Existing studies for solving these topological limitations typically focus only on the convolution of features on network topology, which inevitably relies heavily on network structures. Moreover, most networks are text-rich, so it is important to integrate not only document-level information, but also the local text information which is particularly significant while often ignored by the existing methods. To solve these limitations, we propose \OurGCN, a novel GCN architecture modeling via bidirectional convolution of topology and features on text-rich networks. Specifically, we first transform the original text-rich network into an augmented bi-typed heterogeneous network, capturing both the global document-level information and the local text-sequence information from texts. We then introduce discriminative convolution mechanisms, which performs convolution on this augmented bi-typed network, realizing the convolutions of topology and features altogether in the same system, and learning different contributions of these two parts (i.e., network part and text part), automatically for the given learning objectives. Extensive experiments on text-rich networks demonstrate that our new architecture outperforms the state-of-the-arts by a breakout improvement. Moreover, this architecture can also be applied to several e-commerce search scenes such as JD searching, and experiments on JD dataset show the superiority of the proposed architecture over the baseline methods.
\end{abstract}



\maketitle
{
\renewcommand{\thefootnote}{\fnsymbol{footnote}}
\footnotetext[1]{Both authors contributed equally to this research.}
}
\section{introduction}\label{sec:1 intro}
Real-world systems can often be modeled as networks, including social networks, biological networks and information networks. Network analysis has been an active research topic for decades both in academia and industry. Recently, research on analyzing networks with deep learning has received widespread attention. In particular, graph convolutional networks (GCNs) \cite{19}, which obtain node embeddings through the propagation and aggregation of the features on network topology, have achieved great success and been widely applied in natural language processing \cite{13,14,31}, traffic forecasting \cite{15,16} and recommendation systems \cite{17,18}.

While the success of GCNs and their variants \cite{26,27,28}, a key issue with them is the topological limitations. Take an extreme example, let the number of layers of GCNs be very large, then from the level of methodology, the node features only serve as an initial solution of embeddings, and continuously smooth it based on the topology using the propagation and aggregation mechanisms, making GCNs almost entirely depend on network topology. Specifically, we may get $k$ different embeddings from GCNs if there are $k$ connected components of the network. Especially, when there is only one component of the network (i.e., network is fully connected), all node embeddings may converge to similar values, which is a severe oversmoothing phenomenon. Considering this, GCNs only obtain satisfactory embeddings from a good local mixing state of propagation by limiting the number of propagation to two or three layers. However, this will further make GCNs rely heavily on the local homophily of topology, that is, neighborhoods should be similar, a very strong assumption in many real-world text-rich networks.

There are also some different attempts proposed to design algorithms and models to handle the topological limitations of GCNs. For example, several studies adopt the idea of topology optimization, which refine the network topology by introducing additional schemes such as normalization \cite{1} and markov random fields \cite{2}.
Another line of attempts is the self-supervision based methods, which utilize several highly credible labels derived from GCNs to optimize the topological channels in the following propagation of GCNs \cite{8,3,9}. The third attempt is the skip connection based methods that adaptively select the appropriate neighborhoods for each node from the perspective of jumping knowledge \cite{21,22}. In addition, there are also some other attempts that can be considered to solve the topological limitations of GCNs, such as attention-based methods \cite{4,25}, which leverage the attention mechanisms to allocate appropriate weights to different neighborhoods, so as to refine the network topology. 

These existing methods have achieved reasonable results at handling the topological limitations of GCNs and thus improved the performance of GCNs. But they still focus only on the convolution of features on network topology. However, from the level of model architectures, an ideal way may be that the convolutions of features (on topology) and topology (on features) play together in the same system. The model should also learn correctly the contribution of each part, that is, topology or text, automatically, for given learning objectives. 

Besides, since most real networks are text-rich, it is also important to incorporate not only the global document-level information, but also the local text information that reflects the semantics expressed in natural language, which is particularly significant while often ignored by existing methods.

To solve these above problems, in this paper we propose a novel GCN architecture, i.e., \OurGCN, on text-rich networks. It consists of two main parts, data modeling and discriminative convolution. To be specific, we first augment the original text-rich network into a bi-typed network, as shown in the top-middle part of Fig. 1. It includes two types of nodes, namely the real nodes (e.g., document nodes in the original network) and entity nodes (e.g., phrases or words extracted from text sequences); and three types of edges, that is, edges between real nodes such as paper citations, edges between real nodes and entity nodes which reflect their inclusion relationships, as well as edges between entity nodes which reflect the semantic structure information in text sequences. We then introduce a discriminative 
joint convolution mechanism, based on the concept of meta-path, which can distinguish and learn out the contributions of network part and text part based on the learning objectives. In addition, as the augmented network constructed in data modeling contains both global document-level information and local text information, this architecture itself can incorporate more semantics and knowledge from texts. 

The rest of the paper is organized as follows. Section \ref{sec:2 preliminaries} gives the problem definitions and introduces GCN. Section \ref{sec:3 framework} proposes the new GCN architecture on text-rich networks. Section \ref{sec:4 exp setting} shows the extensive experiment details and the algorithms we compared. In Section \ref{sec:5 exp result and ablation study} we discuss and analyze the experiment results. To extend our method to more fields, Section \ref{sec:6 JD} introduces the application on e-commerce search. Finally, we discuss related work in Section \ref{sec:7 related work} and conclude in Section \ref{sec:8 conclusion}.

\section{Preliminaries}\label{sec:2 preliminaries}
We first introduce the problem definition, and then discuss GCN which serves as the base of our new architecture.
\subsection{Problem Definition}
The notations used in this paper are summarized in Table 1.

\emph{Definition 1.} \textbf{Text-rich Network}. Consider a network $G = (V, E)$, where $V=\left\{v_{1}, v_{2}, \ldots, v_{n}\right\}$ is the set of $n$ nodes and $E = \{e_{ij}\} \subseteq V \times V$ the set of $m$ edges. The topological structure of network $G$ can be represented by an $n\times n$ adjacency matrix $A = (a_{ij})_{n\times n}$, where $a_{ij} = 1$ if there is an edge between nodes $i$ and $j$, or 0 otherwise. Specifically, we call such a network \textbf{text-rich network} if a portion of its nodes is associated with textual information that collectively forms a \emph{corpus $\mathcal{D}$}.

Existing methods (e.g., GCN and GAT) rely heavily on the topology of networks, while placing little emphasis on original texts that endorse high quality information in the network. To make better use of the original texts, we extend the notion of semi-supervised node classification to encompass real-world text-rich networks such as citation networks.

\emph{Definition 2.} \textbf{Semi-supervised Node Classification on Text-Rich network}. Given a text-rich network $G = (V, E)$, a corpus $\mathcal{D}$ and a labeled node set $V_L$ containing $u \ll |V|$ nodes, where each node $v_i \in V_L$ is associated with a label $y_i \in Y$, the objective is to predict labels of $V \verb \ V_L$. Note that each node $v_i \in V$ corresponds to a document $d_i \in \mathcal{D}$.

\begin{table}[htbp]
\caption{The notations used in this paper.}
\begin{tabular}{l|p{5.24cm}}
\hline
Notations & Descriptions\\
\hline
$V, E$ & The sets of nodes and edges\\
\hline
$A, X$& Adjacency matrix and node feature matrix\\
\hline
$G_W = (V_W, E_W)$ & Word network\\
\hline
$G_D = (V_D, E_D)$ & Document network\\
\hline
$G_{W}  = (V_{W'}, E_{W'})$ & Word network with refined edges\\
\hline
$G_{D'} = (V_{D'}, V_{W'})$ & Document network with refined edges\\
\hline
$\mathcal{D}=\{d_1, d_2 ... d_{|V_{D}|}\}$ & The corpus containing documents represented by each document node $v_i \in V_D$\\
\hline
$\T = \{D,W,DW\}$ & Class of sub-network types\\
\hline
$Z$ & Readout result for GCN layers\\
\hline
$H_x^{l}$ & Intermediate feature produced by GCN layer for type $x$ at level $l$\\
\hline
$\mathcal{Y}_D$ & The set of documents that have labels\\
\hline
\end{tabular}
\vspace{-1em}
\end{table}

\begin{figure*}
    \centering
    \includegraphics[width=0.9\linewidth]{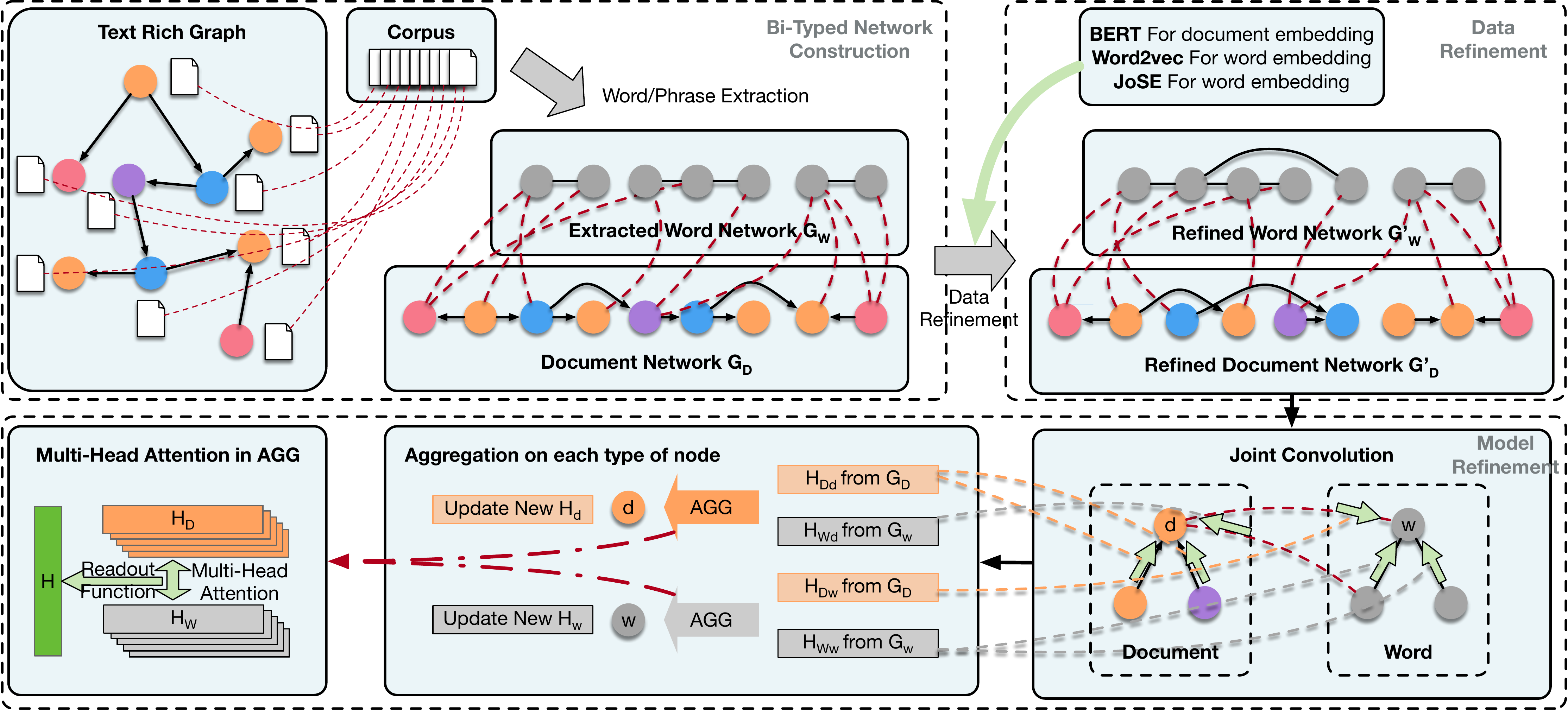}
    \caption{Overview of our \OurGCNBf framework.
    It contains three major parts, bi-typed network construction, joint convolution, as well as data and model refinements.}
    \label{fig:framework}
    \vspace{-1em}
\end{figure*}

\subsection{Graph Convolutional Network}
Graph Convolutional Network (GCN) \cite{19}, originally developed by Kipf and Welling, is a variant of multi-layer convolutional neural networks that operates directly on networks. It learns embeddings of each node by iteratively aggregating the information from its neighborhoods. Mathematically, given a network $G = (V, E)$, let $X\in R^{n\times d}$ be the node feature matrix, where $n=|V|$ denotes the number of nodes and $d$ the dimension of features. Let $A\in R^{n\times n}$ be the adjacency matrix of the network and $D$ the corresponding node degree matrix with $ D_{ii}= \sum\nolimits_{j} A_{ij}$. Assume that every node is connected to itself, i.e., $\Tilde A = A + I$ (where $I$ is the identity matrix). Then, by introducing an effective renormalization trick, that is, $\hat{A}= \Tilde D^{-{1\over2}}\Tilde A \Tilde D^{-{1\over2}}$ (where $\Tilde D_{ii}=  \sum\nolimits_{j}\Tilde A_{ij}$), the classic two-layer GCN can be defined as:
\begin{equation}
Z=f(X,A)= \operatorname{softmax}(\Tilde A \operatorname{ReLU}(\Tilde A X W^{(0)})W^{(1)})
\end{equation}
where $W^{(0)}$ (and $W^{(1)}$) is the weight parameter, ReLU (and softmax) the non-linear activation function, and $Z$ the final output for the assignment of node labels. While GCN works well on several networks analysis tasks such as node classification \cite{29,30}, it still has a fundamental weakness, that is, topological limitations, including over-smoothing and local homophily of topology, which lead to the main contribution in this work: joint convolution in both network structure and features.
\section{The \OurGCNBf Framework}\label{sec:3 framework}
\subsection{Overview}
Previous GCN-based methods only perform convolution within the view of network structure. However, due to the limitations of GCN architecture, the network structure may contain noisy information that prevents the model from learning real knowledge from the network. However, text-rich networks contain significant amount of text information which can guide the convolution process. To leverage the feature-space convolution (i.e., convolution within the corpus), we decompose text-rich networks into bi-typed networks to explicitly perform joint convolution in both node space and feature space. This decomposition introduces a set of word nodes which are compatible with data refinement in semantic meanings.

The \OurGCN framework consists of three major components, namely bi-typed network construction, joint convolution, and data-model refinements. This section will first describe how to construct the bi-typed network based on the text-rich network and then illustrate the detailed convolution technique in the model and finally investigate the refinement methods we adopted in our framework design.

\subsection{Bi-typed Network Construction}\label{subsec:decompose}
As mentioned above, the traditional GCN framework is not suitable for text-rich joint convolution. To best utilize the text-rich information, in this work, we first construct a bi-typed network $G$ based on the given text-rich network $G_D = (V_D, E_D)$. This bi-typed network structure embeds the semantic information and the structure is explainable which is compatible with further data and model refinements. This process can be split into two parts, namely word network construction and whole network completion.
\subsubsection{Word Network Construction}
In text-rich network, each node $v_{i}$ in the network $G_D$ is associated with a feature corpus denoted as $d_{i}$, usually a document describing this node. To best benefit from the corpus, we convert all documents within this corpus as a word sub-network $G_{W}=(V_{W},E_{W})$ where $V_{W}$ is the collection of all representative words or phrases extracted from the corpus $\mathcal{D}$. $E_{W}$ is the edges connecting word pairs $(w_1,w_2)$ chosen from $V_{W}$. The word edges are constructed based on the partial word sharing between different phrases. For example, ``text\_mining'' and ``data\_mining'' are connected since they share the word ``mining''.
\subsubsection{Whole Network Completion}
After the construction of word sub-network, the whole bi-typed network $G$ is completed by adding edges between document nodes $V_{D}$ and word nodes $V_{W}$. A straightforward approach is applying the ``inclusion'' relationship between a document and a word to create edges. We denote the network added above as: $$G_{DW} = (V_{D} \cup V_{W}, E_{DW})$$ where $$E_{DW} = \{(d,w)|w \in d, \forall w \in V_{W}, \forall d \in V_{D}\}$$
Then the whole network is the union of the networks mentioned above: $$G = G_D \cup G_{DW} \cup G_W = (V_D \cup V_W, E_D \cup E_{DW} \cup E_W)$$
\subsection{Data Refinement}\label{subsec:datarefine}
One underlying assumption of GCN based methods is the local homophily of sub-networks. However, this assumption does not apply directly to our case, since edges in both sub-networks ($G_D$ and $G_W$) suffer from several limitations. Edges in $G_D$ are often based on citations, which does not reflect the semantic similarity between documents. As a result, edges may exist among semantically dissimilar objects, while missing among semantically similar objects. Edges in $G_W$ are based on inclusion relationship among phrases extracted from $\mathcal{C}$, which also does not have strong semantic implications, and thus suffer from similarity limitations.

To enhance the semantic feasibility of edges, we perform edge refinement separately on both sub-networks.
\subsubsection{Document Network Refinement}
The edges in document network are citation in the scientific papers. However, due to the limitation of the  author(s)' knowledge, the citation structure may introduce irrelevant or missing edges. The goal of edge refinement is to trim out edges linking semantically distant objects, while adding edges to link semantically close objects. To achieve this, we first need a notion of similarity. For the document network, we use cosine similarity between BERT embedding of documents.

The specific procedure is as follows: we first calculate an embedding for each document with a pre-trained BERT model, and then calculate the pairwise similarities between documents. Document pairs with similarity higher than the upper threshold $T_{D,high}$ is added as new edges, and existing edges with similarity less than a lower threshold $T_{D,low}$ is trimmed off.
\subsubsection{Word Network Refinement}
The initial criteria for finding the edges in the word network is to share partial words\footnote{The words here refers to phrases extracted from the corpus $\mathcal{D}$. Automated Phrase extraction tools such as AutoPhrase\cite{shang_automated_2017} is adopted here. And the phrase version of the word network can better capture the semantic meaning and the relation between phrases.}. e.g., ``text\_mining'' and ``data\_mining'' are connected since they share the word ``mining''. However, chances are that in some cases sharing words do not guaranteed to have correlation in semantic views. For example, for the links between phrases, we use word embedding similarities for the refinements, that is, we remove noise links such as links between “artificial intelligence” and “artificial life” with very low semantic similarities, and add missing links such as links between “LSTM” and “RNN” with very high semantic similarities. To achieve this, we adopt embedding trained on the whole corpus $\mathcal{D}$ to guide the network structure construction, i.e., word network refinement. Here we've used two different embedding methods using both Euclidean space (word2vec \cite{mikolov2013distributed}) and Spherical space (JoSE \cite{meng2019spherical}) to capture the semantic meaning of words. The refinement procedure for the word network is similar to that of the document network, and here we have also two parameters, namely $T_{W,high}$ and $T_{W,low}$ to control the refinement.
\subsection{Joint Convolution}\label{subsec:jointconv}
Different from the previous work that performs message passing in a bi-typed network, we perform joint convolution for both document nodes and word nodes at the same time. A $\OurGCN$ layer consists of two-level message passing operations: 
\begin{enumerate}
    \item GCN within each type of sub-network.
    This is the basic level for our joint convolution, in each sub-network, we perform original GCN convolution to allow message passing within the same type of sub-network.
    \begin{equation}
        H_{t}^{(l+1)}= GCN_{t}(H^{(l)}), \forall t \in \T
    \end{equation}
    In our setting at each layer, there are three different $H_{t}'s$ generated where $t \in \{D,W,DW\}$.
    \item Aggregation among different types of sub-networks.
    After we got the message $H_{t}$ for each type of sub-network, we adopt an aggregation function $AGG$ to merge the message from different types of sub-networks. We would introduce the $AGG$ further in Section \ref{sec:modelrefine}.
    \begin{equation}
        H^{(l+1)}= AGG \bigg( H_{t}^{(l+1)}, \forall t \in \T \bigg)
    \end{equation}
\end{enumerate}
A complete form of two-layer \OurGCN model can be expressed in the following as:
$$Z=AGG_{2}(GCN_{2t}(AGG_{1}(GCN_{1t}{(X)},\forall t \in \mathcal{T})),\forall t \in \mathcal{T})$$
where $X$ is the original features for all nodes in $G$.
Then follow the heuristic loss definition as introduced in GCN or GAT, we define the loss function by using cross entropy as:
\begin{equation}
        \mathcal{L}=- \sum\limits_{ d\in \mathcal{Y}_D}\sum\limits_{f=1}^{F} Y_{df}\operatorname{ln} Z_{df}
\end{equation}
where $\mathcal{Y}_D$ is the set of document indices that have labels, $Y$ the label indicator matrix, and $F$ the dimension of the output features, which is equal to the number of categories.

\subsection{Model Refinement}\label{sec:modelrefine}
As introduced in Section \ref{subsec:jointconv}, an aggregation function $AGG$ is applied at the stage of merging messages passed from different types of sub-networks. For a document node $d$, it receives messages $H_D$ from GCN module in document sub-network $G_D$ and $H_W$ from GCN module in word sub-network $G_W$. Naive readout functions such as MEAN and CONCAT suffer from not capturing the relation between different sources of the messages. To best utilize the information from different sources of the messages and learn the relation and the interaction between them, we adopt a multi-head attention layer to learn how the messages from document sub-network and word sub-network are used in the final task. For a single-head attention module, it takes both $H_{Dd}$ from document sub-network and $H_{Wd}$ from word sub-network. Then a non-linear function $\rho$ is applied to the concatenation of the two messages to represent the attention for the two messages.
\begin{equation}
    \mathbf{a_d} = \rho (H_{Dd}\|H_{Wd})
\end{equation}
Then the final single-head representation for document node $d$ is calculated by: 
\begin{equation}
    H_{d,single} = \mathbf{a_d} \begin{bmatrix}
H_{Dd}\\
H_{Wd}
\end{bmatrix}
\end{equation}
Following the multi-head attention's intuition, the final representation for document node $d$ can then be calculated by concatenating the outputs of all attention heads:
\begin{equation}
    H_{d} = \bigg|\bigg| H_{d,single}
\end{equation}

\subsection{The Generality of the Framework}
The classical GCN model \cite{19} belongs to transductive learning which limits its scalability. In network representation, different from transductive learning's matrix-by-matrix working mechanism, the inductive learning (such as GraghSage \cite{23} and GAT \cite{4}) applies node-by-node working mechanism, making it generalizable to the unseen nodes. To distinguish these two different learning styles, we give the following formalization. We use $V'_{D}$, $V'_{W}$, $G'_D$, $G'_W$ and $G'_{DW}$ to denote the document nodes, word nodes, document sub-network, word sub-network and document-word sub-network in the testing network $G'$. Then $G'$ can be formalized as: $$G' = G'_D \cup G'_{DW} \cup G'_W \quad \mathrm{and}\quad  G'_{DW} = (V'_{D} \cup V'_{W}, E'_{DW})$$ where $$E'_{DW} = \{(d',w')|w' \in d', \forall w' \in V'_{W}, \forall d' \in V'_{D} \}$$
The \OurGCNBf framework is compatible with the transductive learning and inductive learning, so we develop both versions for it. In the inductive learning version, the node-by-node convolution calculation order needs to be specifically designed.
\section{Experiment setting}\label{sec:4 exp setting}
In this section, we conduct extensive experiments to evaluate the effectiveness of our proposed method \OurGCN compared to several state-of-the-arts on four text-rich networks.
\subsection{Datasets}
We examine our \OurGCN on four text-rich citation networks. Table \ref{tab:dataset} shows the statistics for these datasets.

\begin{enumerate}
    \item \textbf{Cora-Enrich}\footnote{http://zhang18f.myweb.cs.uwindsor.ca/datasets/}: Cora dataset contains 2,708 scientific publications in machine learning, connected by 5,429 citation links. Each paper is manually labeled as one of seven categories: ``Case Based'', ``Genetic Algorithms'', ``Neural Networks'', ``Probabilistic Methods'', ``Reinforcement Learning'', ``Rule Learning'', and ``Theory''. This dataset shares the same citation information with the standard cora dataset while contains whole corpus instead of feature vectors \cite{24}.
    \item \textbf{DBLP-Five}\footnote{https://www.cs.cornell.edu/projects/kddcup/datasets.html.\label{data}} : This is a sub-collection of the original DBLP datasets containing 6,936 scientific papers in five sub-area in computer science: ``High-Performance Computing'', ``Software engineering'', ``Computer networks'', ``Theoretical computer science'', and ``Computer graphics: Multimedia''. These papers are connected by 12,353 links.
    \item \textbf{Hep-Small}\textsuperscript{\ref {data}} and \textbf{Hep-Large}\textsuperscript{\ref {data}}: Two text-rich citation datasets about scientific papers in physics. Hep-Small contains 397 scientific papers in three categories: ``Nucl.Phys.Proc.Suppl'', ``Phys.Rev.Lett'', and ``Commun.Math.Phys'', connected by 812 links. Hep-Large contains 11,752 documents in four categories: ``Phys.Rev'', ``Phys.Lett'', ``Nucl.Phys'', and ``JHEP'', connected by 134,956 links. 
\end{enumerate}
\vspace{-1em}
\begin{table}[htbp]
    \centering
    \caption{Dataset Statistics. $|\mathcal{N}|$ and $|\mathcal{E}|$ are the number of nodes and edges in the citation network. $|\mathcal{C}|$ is the the number of categories.}
    \vspace{-0.5em}
    \begin{tabular}{ cccc }
    \toprule
    \textbf{Datasets} & {$|\mathcal{N}|$} & $|\mathcal{E}|$ & $|\mathcal{C}|$\\
    \midrule
    \textbf{Cora-Enrich} & 2,708 & 5,429 & 7\\
    \textbf{DBLP-Five} & 6,936 & 12,353 & 5\\
    \textbf{Hep-Small} & 397 & 812 & 3\\
    \textbf{Hep-Large} & 11,752 & 134,956 & 4\\
    \bottomrule
    \end{tabular}
    \label{tab:dataset}
    \vspace{-1em}
\end{table}
\subsection{Compared Algorithms}\label{subsec:comparealgm}
We compare our methods and its variants with four state-of-the-art methods, introduced as follows:
\begin{itemize}
    \item \textbf{GCN} \cite{19}: It is a classical graph convolution network model and the main baseline of our \OurGCN. Specifically, the features here are extracted directly from corpus $\mathcal{D}$.
    \item \textbf{GAT} \cite{4}: Based on GCN model, it introduces the node-level multi-head attention mechanisms to specify the weights from different neighborhoods. 
    \item \textbf{DGI} \cite{velickovic2019deep}: It is an unsupervised graph neural network model which derives node embeddings by maximizing mutual information between patch representations and corresponding high-level summaries of graphs.
    \item \textbf{Geom-GCN} \cite{7}: It is a semi-supervised graph neural network model that obtains node embeddings by adopting a bi-level aggregator operating on the structural neighbors.
    \item \textbf{\OurGCNBf-B}ase: The basic model of \OurGCN, (i.e., without any refinement).
    \item{\textbf{\OurGCNBf-R}efine}: \OurGCN of adding data refinement.
    \item \textbf{\OurGCNBf-A}ttention: \OurGCN of adding model refinement (attention).
    \item{\textbf{\OurGCNBf-R}efine-\textbf{A}ttention}: 
     \OurGCN of adding both data refinement and model refinement (attention).
\end{itemize}
\begin{table*}[htbp]
    \centering
    \caption{Overall results (Accuracy) on four mentioned datasets.}
    \vspace{-1em}
    \begin{tabular}[width=\linewidth]{ ccccc }
    \toprule[0.5pt]
    \toprule[0.5pt]
    \textbf{Methods} & \textbf{Cora-Enrich}  & \textbf{DBLP-Five} & \textbf{Hep-small} & \textbf{Hep-large}  \\
    \midrule[0.5pt]
    GCN& 0.8815 &0.9177&0.4872&0.5243\\
    GAT& 0.8852 &0.9365&0.4915&0.5270\\
    DGI& 0.8918 &0.8564&0.5108&---\\
    GeomGCN& 0.8926 &0.9307&0.4872&0.5047\\
    \midrule[0.5pt]
    \OurGCN-B& 0.9000 &0.9336&0.5359&0.5277\\
    \OurGCN-R& 0.9148 &0.9466&0.5897&0.5345\\
    \OurGCN-A& 0.9222 &0.9380&0.5385&0.5260\\
    \OurGCN-R-A& \textbf{0.9370} &\textbf{0.9524}&\textbf{0.6667}&\textbf{0.5489}\\
    \bottomrule[0.5pt]
    \bottomrule[0.5pt]
    \end{tabular}
    \vspace{-1em}
    \label{tab:result}
\end{table*}

\subsection{Hyper-Parameter and Evaluation}
For all GCN based methods, a two-layer architecture is used to avoid oversmoothing and achieve better performance. For data refinement, $T_{D,high}$ and $T_{W,high}$ are set to 0.95 to include only the high quality and correlated node pairs into the refined network, $T_{D,low}$ and $T_{W,low}$ are set to 0.5 to filter out unrelated node pairs. For attention-based methods, the number of heads is set to 4 for fair comparisons. All other parameters for baselines methods are set to their default values. We evaluate all models using node classification accuracy. 
\section{Experiments Results and Ablation Study}\label{sec:5 exp result and ablation study}

Table \ref{tab:result} shows the overall results on the node classification task performed on the datasets mentioned above. 

\subsection{Comparisons to Existing Methods}
As shown in Table 3, our model \OurGCN outperforms the basic GCN by a large margin, which demonstrates the superiority of our new bidirectional convolution mechanisms of topology and features. It also performs much better than the other state-of-the-art GCN methods, which further validate the effectiveness of our new architecture. Note that as illustrated in the model design part, our approach is almost orthogonal to many existing GCN models which can incorporate our strategy easily to further improve performance.
\vspace{-1em}
\begin{figure}[htbp]
\centering
\setlength{\abovecaptionskip}{-0.3cm}
\setlength{\belowcaptionskip}{0cm}
\includegraphics[width=0.48\textwidth]{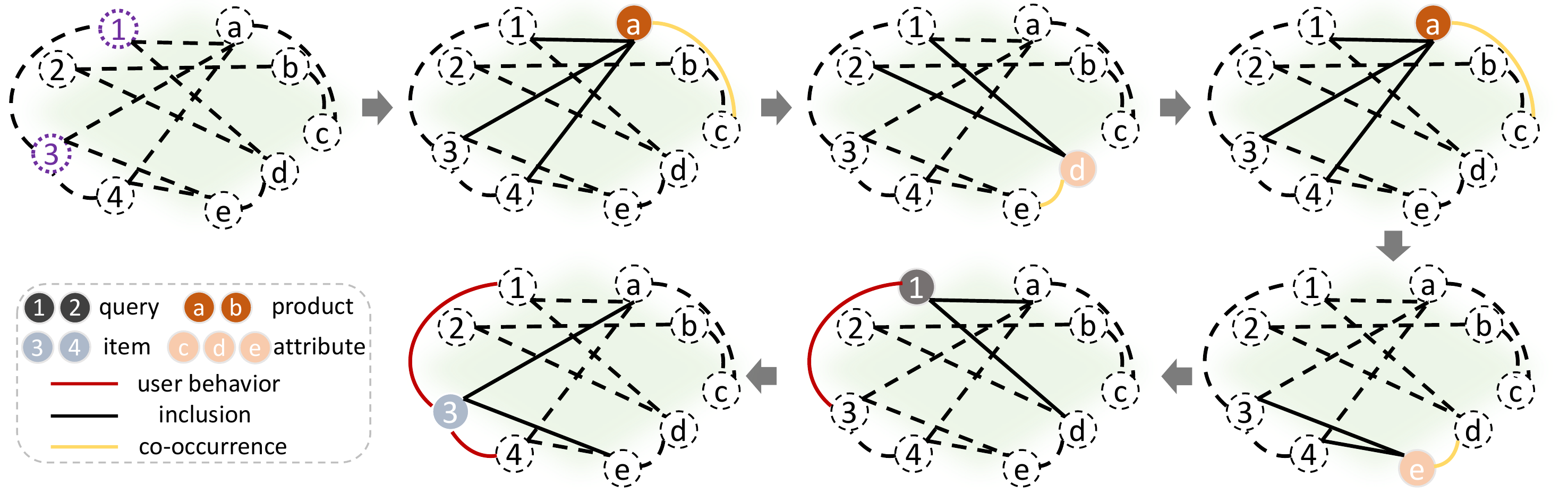}
\hspace{1in}
\caption{The convolution process in \OurGCNBf for a real e-commerce network. The target nodes are query node \#1 and item node \#3, so we orderly update \#1's neighbor nodes (product node \#a and attribute node \#d), \#3's neighbor nodes (product node \#a and attribute node \#e), and finally \#1 and \#3 themselves.}
\label{conv_order}
\vspace{-2em}
\end{figure}
\subsection{Refinement and Attention}
To further clarify our claimed contribution, we compare 
\OurGCN with several variations as stated in Section \ref{subsec:comparealgm}. We claim that the data refinement together with the model refinement make a great contribution on boosting the performance.
\subsubsection{Data Refinement}
Compare our \OurGCN-B's result with original GCN, our proposed bi-typed convolution architecture can capture the semantic meanings embedded in the network structure. Then adding refinement can further increase the performance of \OurGCN-B.
The result comparison between \OurGCN-B and \OurGCN-R shows that the semantic (data) refinement can benefit the base \OurGCN model. 
\subsubsection{Model Refinement (Attention)}
The model refinement (attention) introduces another performance boost. In Table \ref{tab:result}, the comparison of two model pairs (\OurGCN-B, \OurGCN-A) and (\OurGCN-R, \OurGCN-R-A) shows the power of this attention refinement. It performs as effective as data semantic refinement because the attention mechanism can capture the information on how the message from different types of the network is utilized. However, in DBLP and Hep datasets, the attention itself cannot bring significant performance boost compared with data refinement. This is because data refinement can capture more semantic meaning in the original text-rich network; however, attention only enhances the original network structure information, which itself is not enough to solve the problem independently. After the data refinement, the model is able to capture more semantic information. At that time, the attention can show its power to learn the different weights or utilization of different message from each type of network.

\begin{figure*}[htbp]
\centering
\vspace{-1em}
\includegraphics[width=0.8\textwidth]{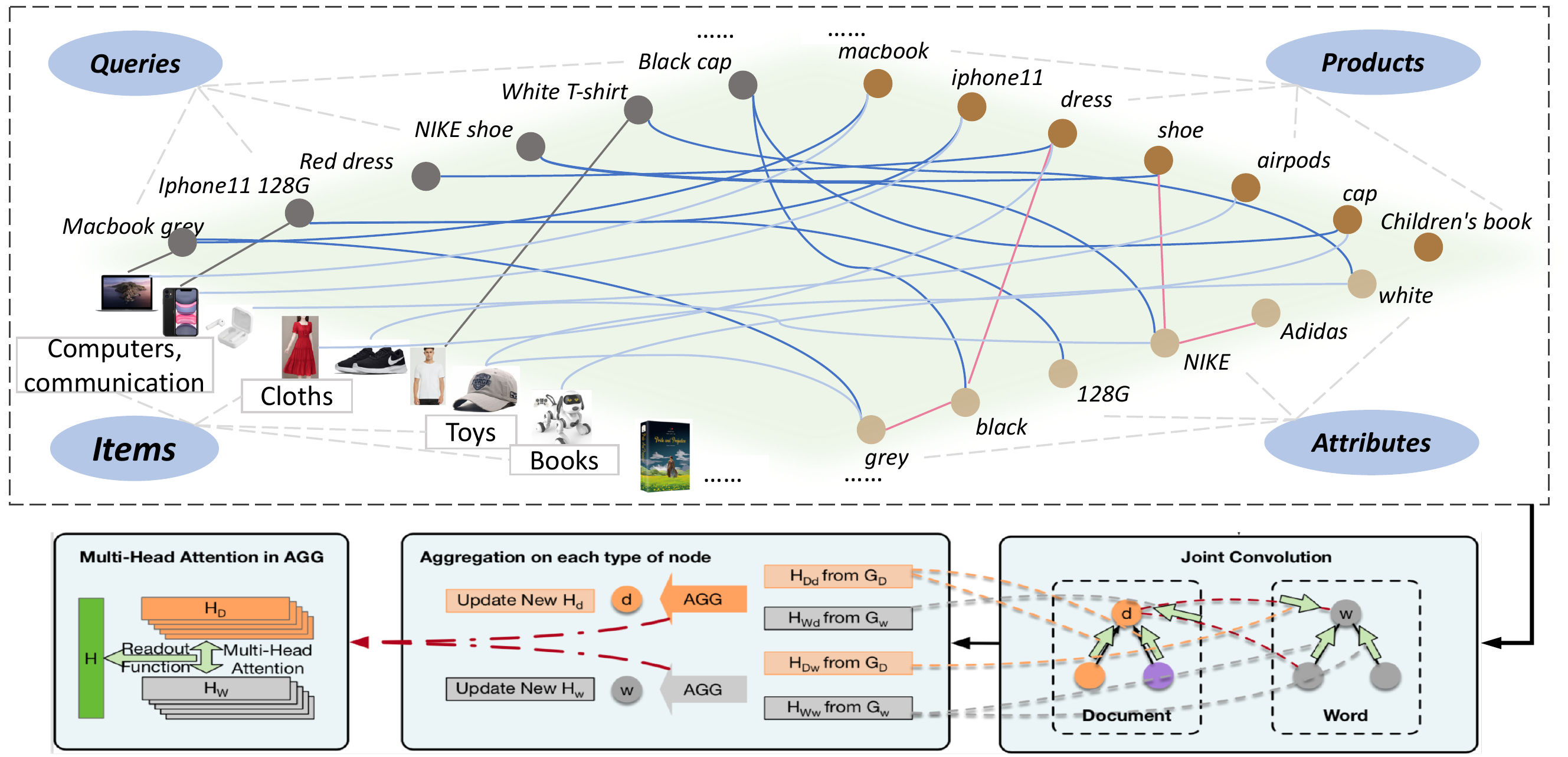}
\caption{An example of the augmented network in a real e-commerce search scene. The edges between two real nodes, two entity nodes, one real node and one entity node are generated via the relationships of user behavior, co-occurrence and inclusion, respectively.}
\label{JD}
\vspace{-1em}
\end{figure*}
\vspace{-.5em}
\section{Application on E-commerce Search}\label{sec:6 JD}
\textbf{E-commerce network construction.} Our new framework is also correctly suitable for the search scene in some e-commerce platforms. To be specific, in an e-commerce search scene, the queries and items which are from users and online shops respectively could constitute a natural bi-typed heterogeneous network (e.g., in Figure \ref{conv_order}). Learning node embedding from this network is helpful to give an accurate estimation of relevance between each pair of query and item, such that it can optimize the follow-up ranking results. From JD.com, we build such a network where a total of 6.5M queries and 50M items covering all categories existed as the real nodes. In consideration of the product and attribute phrases playing important roles in the e-commerce corpus, we extract 60K product phrases and 12K attribute phrases from JD's database to act as the entity nodes in the network. Some representative examples from this built network are shown in Fig. \ref{JD}. The red edge connecting a pair of real nodes generated via integrating the original user purchase behaviors and BERT-refined user click behaviors, while the yellow edge from an entity node to another is selected via retrieving the most co-occurrence phrase.

\textbf{Baseline algorithms.} We compare our models with two types of baseline algorithms which are classical deep text matching models and emerging graph neural network (GNN) models for the searching problem. The former usually adopts a representation-based or interaction-based matching pattern to calculate a relevance score between two pieces of texts. The latter leverages neural network architectures (i.e., graph convolution layer with or without graph attention layer), to deal with network-structure data.. 
A brief introduction to these related methods and \OurGCN models is proviuded as below:
\begin{itemize}
\item{\textbf{MV-LSTM} \cite{mv-lstm}}: It is a representation-based method, which uses Bi-LSTM architecture to learn two sentences' separate embedding with contextual information.
\item{\textbf{ARC-I} \cite{arc}}: It is a representation-based method, which captures two sentences' different-level matching signals with the Siamese CNN architecture.
\item{\textbf{ARC-II} \cite{arc}}: As an enhanced version of ARC-I, it is interaction-based, and directly models the interaction space of two sentences with a single group of CNNs, which is different from the Siamese architecture in ARC-I.
\item{\textbf{K-NRM} \cite{k-nrm}}: It is an interaction-based method, using the kernel-based technique to learn multi-level matching signals.
\item{\textbf{MatchPyramid} \cite{matchpy}}: It underlines the similarity of text match and image recognition, and thus designs a hierarchical CNN's structure to deal with the text match problem.
\item{\textbf{DUET} \cite{duet}}: It is a combination method of the representation-based model and interaction-based model.
\item{\textbf{HG4SM} \cite{hg4sm}}: It contains HG4SM-1 (representation-based embedding), HG4SM-2 (interaction-based embedding) and HG4SM-3 (metapath-guided heterogeneous network embedding) to solve the relevance match problem. We compare our model with all its three versions.
\item{\textbf{GCN} \cite{19}, \textbf{GAT} \cite{4}, \textbf{\OurGCNBf-R}efine, \textbf{\OurGCNBf-R}efine-\textbf{A}ttention}: As introduced in Section 4.2.
\end{itemize}
\textbf{Metrics}. Six kinds of metrics are introduced to fully evaluate the above models’ performance. These metrics are widely used in machine learning, including AUC (Area Under the receiver operating characteristic Curve), Accuracy, Precision, Recall, F1-score and FNR (False Negative Rate). It is worth mentioning that we mainly focus on the result of AUC in the JD search relevance scene.

\textbf{Experimental setup and result analysis}. For simplicity, each real or entity node’s embedding is achieved by calculating the average of word embeddings, which are initialized by the pre-trained results on 2 billion e-commerce corpus. We use the Lazy-Adam optimizer with a learning rate of 1.0e-3. For \OurGCN-R-A model, we set 4-head attention calculation for the aggregation of node embedding and meta-path embedding. Based on 39K human-labeled evaluation dataset, we select the best model and report its results in Table \ref{app-results} corresponding to the maximum AUC value within 3 epochs. 

For the results, on the one hand, the GNN-type methods are superior to all of compared deep text match method such as MV-LSTM and MatchPyramid, which benefits by the extra refinement of the rich local neighbor information. On the other hand, by disentangling the real nodes and entity nodes from the heterogeneous network and afterward building up two convolution operations on homogeneous nodes and heterogeneous nodes, our models \OurGCN-R and \OurGCN-R-A also outperform the two classical GNN models (i.e., GCN and GAT) in all the test sets.

\begin{table*}[ht]
\caption{The performance of deep text match models, GNN models and our models. (*) is the dominant metric used in JD.}
\begin{center}
\vspace{-1em}
\begin{tabular}{cccccccc}
\toprule[0.5pt]
\toprule[0.5pt]
Models &AUC(*) &Accuracy &Precision & Recall &F1-score &FNR \\
\midrule[0.5pt]
MV-LSTM &0.8278 &0.8023 &0.8021 &0.9873 &0.8851 &0.8224 \\
K-NRM &0.8021 &0.7918 &0.7942 &0.9854 &0.8796 &0.8618\\
ARC-I &0.7345 &0.7771 &0.7769 &0.9975 &0.8735 &0.9669\\
ARC-II &0.7783 &0.7920 &0.7915 &0.9915 &0.8803 &0.8815\\
MatchPyramid &0.8007 &0.7946 &0.7988 &0.9806 &0.8805 &0.8336\\
DUET &0.8077 &0.7799 &0.7789 &\textbf{0.9980} &0.8749 &0.9563\\
HG4SM-1 &0.8067 &0.7978 &\textbf{0.8855} &0.8474 &0.8660 &\textbf{0.3697}\\
HG4SM-2 &0.8289 &0.8534 &0.8770 &0.9421 &0.9084 &0.4459\\
\midrule[0.5pt]
HG4SM-3 &0.8500 &0.8592 &0.8852 &0.9392 &0.9114 &0.4110\\
GCN &0.8458 &0.8580 &0.8816 &0.9425 &0.9110 &0.4272\\
GAT &0.8523	&0.8539	&0.8819	&0.9361	&0.9082 &0.4234\\
\midrule[0.5pt]
\OurGCN-R &0.8564 &0.8598 &0.8847 &0.9409 &0.9119 &0.4139\\
\OurGCN-R-A &\textbf{0.8635} &\textbf{0.8601} &0.8850 &0.9409 &\textbf{0.9121}  &0.4129\\
\bottomrule[0.5pt]
\bottomrule[0.5pt]
\end{tabular}
\end{center}
\label{app-results}
\vspace{-1em}
\end{table*}
\section{Related Work}\label{sec:7 related work}
Graph convolutional networks (GCNs) aim to extract high-level features from nodes and their neighborhoods by using the propagation and aggregation mechanisms. However, a key issue with GCNs is their topological limitations (i.e., over-smoothing and local homophily of topology). Existing methods for solving these topological limitations can be mainly divided into four categories, topology optimization methods, self-supervised methods, skip connection methods and attention-based methods. 

\textbf{Topology Optimization Methods.} There are several studies that adopt the idea of topology optimization to improve GCNs. For example, DropEdge \cite{5} proposes to reduce the message passing by randomly deleting a certain number of edges from the input network in order to alleviate the topological limitations of GCNs. MRFasGCN \cite{2} makes a community knowledge-based MRF as a layer of convolution of GCN to relieve its topological limitations. TO-GCN \cite{6} refines network topology by employing the given labels as the pairwise constraints (must-link and cannot-link). Geom-GCN \cite{7} utilizes the information (discriminative structures and long-range dependencies) in an embedding space from a network and a bi-level aggregation scheme to relieve the impact of network topology on GCNs performance.

\textbf{Self-supervised Methods.} Another distinct line for overcoming the topological limitations of GCNs is to augment the original label set by adding the high-credible labels derived from GCNs. For example, Li \emph{et al.} \cite{8} propose a self-training approach, which first trains GCN with given labels, and then adds the most confident predictions for each class to label set for subsequent training of GCN. Self-enhanced GNN \cite{3} employs two algorithms, topology update and training node augmentation, to improve the quality of input data to promote the performance of GCNs. M3S \cite{9} proposes a multi-stage training algorithm on the basis of self-training, which first adds confident data with virtual labels to the label set, and then applies DeepCluster on the embedding process of GCN and constructs a novel self-checking mechanism to improve the training of GCN. DSGCN \cite{10} extends M3S by adopting a threshold-based rule, that is, insert an unlabeled node if and only if its classification margin is above the threshold, to augment the training set to ease the topological limitations of GCNs.

\textbf{Skip Connection Methods.} The third strategy utilizes the idea of skip connection that adaptively selects appropriate neighborhoods for each node, to overcome the topological limitations of GCNs. For instance, JKNet \cite{22} introduces jumping knowledge networks, which flexibly leverages different neighborhood ranges for each node, to enable better structure-aware representation. Fey \cite{21} improves JKNet by exploring a highly dynamic neighborhood aggregation procedure that aggregates neighborhood representations of different localities.

\textbf{Attention-based Methods.} Some work using the attention mechanisms to allocate appropriate weights to different neighborhoods in network can also be considered to solve the topological limitations of GCNs. For example, GAT \cite{4} introduces the attention mechanisms, which uses weights on links, to aggregate the information of neighborhoods to refine network topology. Liu \emph{et al}. \cite{11} propose a non-local aggregation framework with an efficient attention-guided sorting to put the distant but informative nodes near each other, so as to relieve the topological limitations of GCNs. AGNN \cite{12} adopts the method that removes all the intermediate fully-connected layers, and replaces the propagation layers with attention mechanisms which learn a dynamic and adaptive local summary of the neighborhood to achieve more accurate predictions. SPAGAN \cite{20} designs a path-based attention that explicitly considers the influence of a sequence of nodes yielding the minimum cost, or shortest path, between nodes and its neighborhoods, mitigating the topology limitations of GCNs.

Though those methods improve the performance of GCNs to some extent, they still have several essential limitations. That is, the convolutions of these methods mainly focus on using features for network topology, making them heavily dependent on network structure. At the same time, most networks are text-rich, it is important to integrate not only the global document-level information, but also the local text-sequence information which contains important semantics while is often ignored by the existing methods.

\textbf{Text GCN.} Also of note, a Text-GCN\cite{yao2019graph} has been proposed very recently to extend the network into words and put the whole system under GCN architecture. However, they did not change the architecture of GCN. Neither the word sequence semantic information nor the heterogeneity underlined in the Bi-Typed network was considered.
\section{conclusion}\label{sec:8 conclusion}
We propose a new GCN architecture, namely \OurGCN, for text-rich networks, to overcome the topological limitations of GCNs. This is the first time to relieve these topological limitations, including over-smoothing and local homophily of topology, through the convolutions of network and text in the same system. Meanwhile, by utilizing a discriminative hierarchical convolution mechanism, based on the concept of meta-path, we can learn the contributions of these two parts, that is, network part and text part, automatically aiming to the ground truth such as node classification. In addition, we incorporate more semantic and knowledge information from texts, not only the global document-level information, but also the local text-sequence level information, together to convolve, making the model more powerful. 

Empirical results on several text-rich networks demonstrate that our new architecture has a breakout improvement
over the state-of-the-arts. 
Meanwhile, our new architecture is also well applied to some e-commerce search scenes (e.g., JD searching). Last but not least, this architecture is almost orthogonal to many existing GCN methods and thus can be readily incorporated to further improve their performance.
\section{Acknowledgements}
Research was sponsored in part by US DARPA KAIROS Program No. FA8750-19-2-1004 and SocialSim Program No.  W911NF-17-C-0099, National Natural Science Foundation of China No. 61772361, National Science Foundation IIS-19-56151, IIS-17-41317, IIS 17-04532, and IIS 16-18481, and DTRA HDTRA11810026.
\bibliographystyle{ACM-Reference-Format}
\bibliography{acmart}
\end{document}